\pdfoutput=1

\documentclass[11pt]{article}

\usepackage[final]{acl}

\usepackage{times}
\usepackage{latexsym}

\usepackage[T1]{fontenc}

\usepackage[utf8]{inputenc}

\usepackage{microtype}

\usepackage{inconsolata}

\usepackage{graphicx}
\usepackage{amsmath}
\usepackage{lipsum}
\usepackage{amssymb}
\usepackage{enumitem}
\usepackage{multirow}
\usepackage{adjustbox}
\usepackage{multicol}
\usepackage{arydshln}
\usepackage{booktabs}
\usepackage{hyperref}
\title{MultiPL-MoE: Multi-Programming-Lingual Extension of Large Language Models through Hybrid Mixture-of-Experts}

\author{\small{Qing Wang, Xue Han \thanks{The corresponding author.}, Jiahui Wang, Lehao Xing, Qian Hu, Lianlian Zhang, Chao Deng, Junlan Feng \footnotemark[1]} \\
  \small{JIUTIAN Team, China Mobile Research Institute, Beijing, China} \\
  \texttt{\scriptsize{\{wangqingai, hanxueai, wangjiahui, xinglehao, huqianai, zhanglianlian, dengchao, fengjunlan\}@chinamobile.com}} \\}
  
\begin{document}
\maketitle
\begin{abstract}
Despite LLMs' excellent code creation capabilities, multilingual code generation remains extremely challenging. To address this, we intent to improve the multi-programming-lingual (MultiPL) performance of the base LLMs while retaining the most popular ones using restricted computational resources. We consider MultiPL to be a special case of multiple natural languages and propose a MultiPL extension of LLMs utilizing a hybrid mixture of experts (MoE), called MultiPL-MoE. Specifically, MultiPL-MoE combines two paired MoEs to optimize expert selection at both the token and segment levels. The \textbf{token-level MoE} is a standard upcycling MoE structure with a shared expert and a novel gate weight normalization approach that aids in the final fusion with the segment-level MoE. The \textbf{segment-level MoE} incorporates two innovative designs to better capture the syntactic structure and contextual patterns of programming languages: First, using a sliding window to partition the input token sequence into multiple segments; Then, adopting an expert-choice routing strategy that allows experts to select the top-k segments. The results of the experiment proved the effectiveness of MultiPL-MoE. \footnote{\url{https://github.com/Eduwad/MultiPL-MoE}}.

\end{abstract}

\section{Introduction}
Large Language Models (LLMs) such as GPT4 \citep{gpt4} and Qwen2.5-coder \citep{qwen2.5-coder-1.5B} have shown remarkable capabilities in code generation, facilitating software engineers to boost their efficiency. However, many studies have highlighted a significant discrepancy between performance on high-resource programming languages such as Python and low-resource ones like Rust \citep{survey-code}, which refers to those with limited availability both online and in training datasets.

\begin{figure}[t]
  \includegraphics[width=\columnwidth]{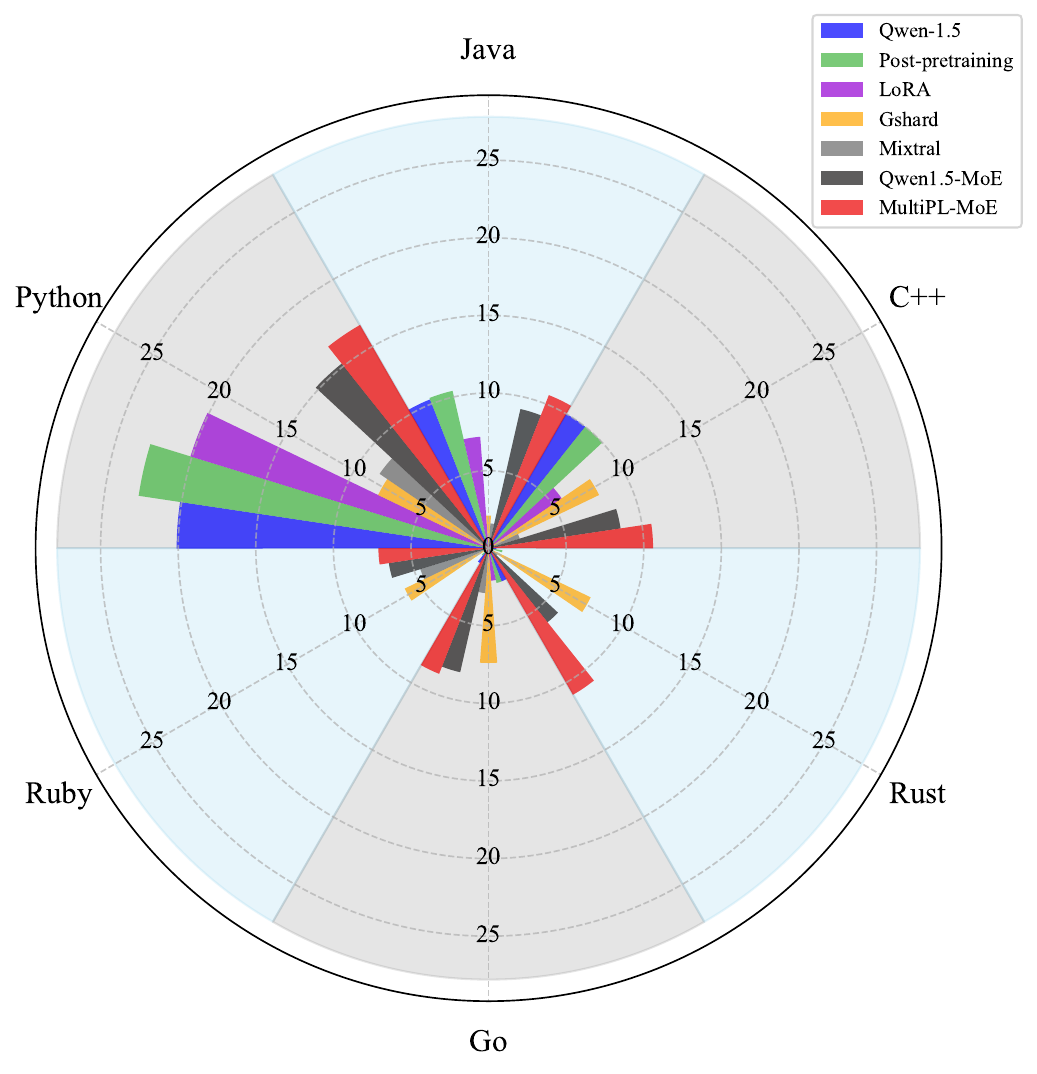}
  \caption{MultiPL-MoE achieves dual success on HumanEval by preserving proficiency in high-resource programming languages and advancing performance on low-resource languages. Specially, it attains a pass@1 score surpassing the second-best model by 1.97 points on low-resource languages while sacrificing only 0.9 points below Qwen1.5 on high-resource languages.}
  \label{fig:code-polar}
\end{figure}

The growing demand for multilingual code generation capabilities has driven efforts to inject extensive programming language knowledge into LLMs. To enhance the base LLMs’ multi-programming-lingual (\textbf{MultiPL}) capabilities, some existing methods \citep{survey-183, survey-75, survey-76, survey-145, han2023survey, survey-63, loire} continue train a base LLM with data from multiple programming languages but are highly computationally intensive. Others \citep{survey-130, survey-38, survey-57, survey-180, survey-188, survey-195} improve specific programming languages by fine-tuning a base LLM with specific high-quality low-resource data. However, they suffer from severe catastrophic forgetting for the base model’s existing code capabilities.

Facing these challenges, we intend to improve LLMs' MultiPL performance, especially for low-resource programming languages, while maintaining the performance of existing popular ones with a relatively small additional computational budget. Recent advances in programming language translation have demonstrated that they share common semantics \citep{khan-etal-2024-codeeval} and can be converted from one to another in the same way that natural languages do \citep{semi-code}. Consequently, we may enhance the MultiPL performance of LLMs by leveraging recent advancements in natural language expansion technology, which adopt Mixture-of-Experts (MoE) architecture by upcycling methods and prove to achieve good language expansion effectively while preventing catastrophic forgetting \citep{moe-lpr}.


In this paper, we take MultiPL as a special case of multiple natural language expansion and propose MultiPL-MoE, a multi-programming-lingual (MultiPL) extension of LLMs through hybrid mixture-of-experts (MoE). Unlike prior MoE methods for multiple natural language extension \cite{moe-lpr} that focus on internal language consistency by predicting subsequent tokens but ignore the internal syntax structures of programming languages \citep{coderosetta}, MultiPL-MoE employs novel hybrid MoE architecture for fine-grained learning of both the token-level semantic features and segment-level syntax features, enabling the model to reason about code structure by identifying and categorizing different syntactic elements. 

The token-level MoE combines traditional token-choice routing with a shared expert \cite{deepseekmoe, ding2024mathcal} and a novel routing weight normalization mechanism to handle scale mismatch during the later fusion with the segment-level MoE.

In particular, MultiPL-MoE combines two paired MoEs to optimize expert selection at both the token and segment levels. The token-level MoE is a typical upcycling MoE structure with a shared expert \citep{deepseekmoe, ding2024mathcal, mocle} and a novel gate weight normalization approach that aids in the final fusion with the segment-level MoE. To better capture the syntactic structure and contextual patterns of programming languages, the segment-level MoE incorporates two innovative designs: 1) Using a sliding window to partition the input token sequence into multiple segments. 2) Adopting a different routing strategy with segment-level MoE that allows experts to select the top-k segments instead of segments selecting experts.

As illustrated in Figure \ref{fig:code-polar}, experiment results show that MultiPL-MoE achieves balanced MultiPL proficiency with 1.97\% higher average accuracy than baselines on low-resource languages (Rust, Go, Ruby) in the HumanEval benchmark, while matching state-of-the-art performance within a 1.80\% gap for high-resource languages like Python, Java, and C++.

\section{Related Work}
\textbf{Multilingual code LLM.} Recent work has developed specialized language models for technical domains, including PolyCoder \citep{survey-183} (multi-language code generation), Hardware Phi-1.5B \citep{survey-75} (hardware domain), SketchGen \citep{survey-76} (CAD sketch), Ansible Wisdom \citep{survey-145} (YAML automation), and adversarial PowerShell code generation \citep{survey-63}, which are highly computationally intensive for domain-specific adaptations. Other studies have addressed programming language challenges through diverse approaches, including multilingual model finetuning \citep{survey-130}, code-to-code augmentation for data-scarce languages \citep{survey-38}), and verifiable PLC code synthesis in industrial contexts \cite{survey-57}.

\textbf{Mixture-of-Experts (MoE).} Scaling large language models (LLMs) is critical to enhance linguistic capabilities. MoE can achieve scalable capacity expansion with sub-linear computation costs \citep{shazeer}, which replaces dense FFN layers with multiple experts and a router. GShard \citep{gshard}, Switch Transformer \citep{switch}, ST-MoE \citep{st-moe}, and GLaM \citep{glam} require full training from scratch. Sparse upcycling \citep{sparse} offers a cost-effective alternative by converting pretrained dense models into sparsely activated MoE models (e.g., Skywork-MoE \citep{skywork-moe}, LlaMA-MoE \citep{llama-moe}). Several works mainly focused on establishing adaptive routing mechanisms to optimize expert allocation. \citet{shazeer} introduces top-k experts routing with auxiliary loss. Expert choice routing \citep{expert-choice} independently selects the top-k tokens with regularization mechanism. However, the above work simultaneously redundant knowledge integration. Recent work \citep{deepseekmoe, openmoe, qwen_moe, mocle} proposes always-activated shared experts to address this limitation.


\begin{figure*}[t]
  \includegraphics[width=\textwidth]{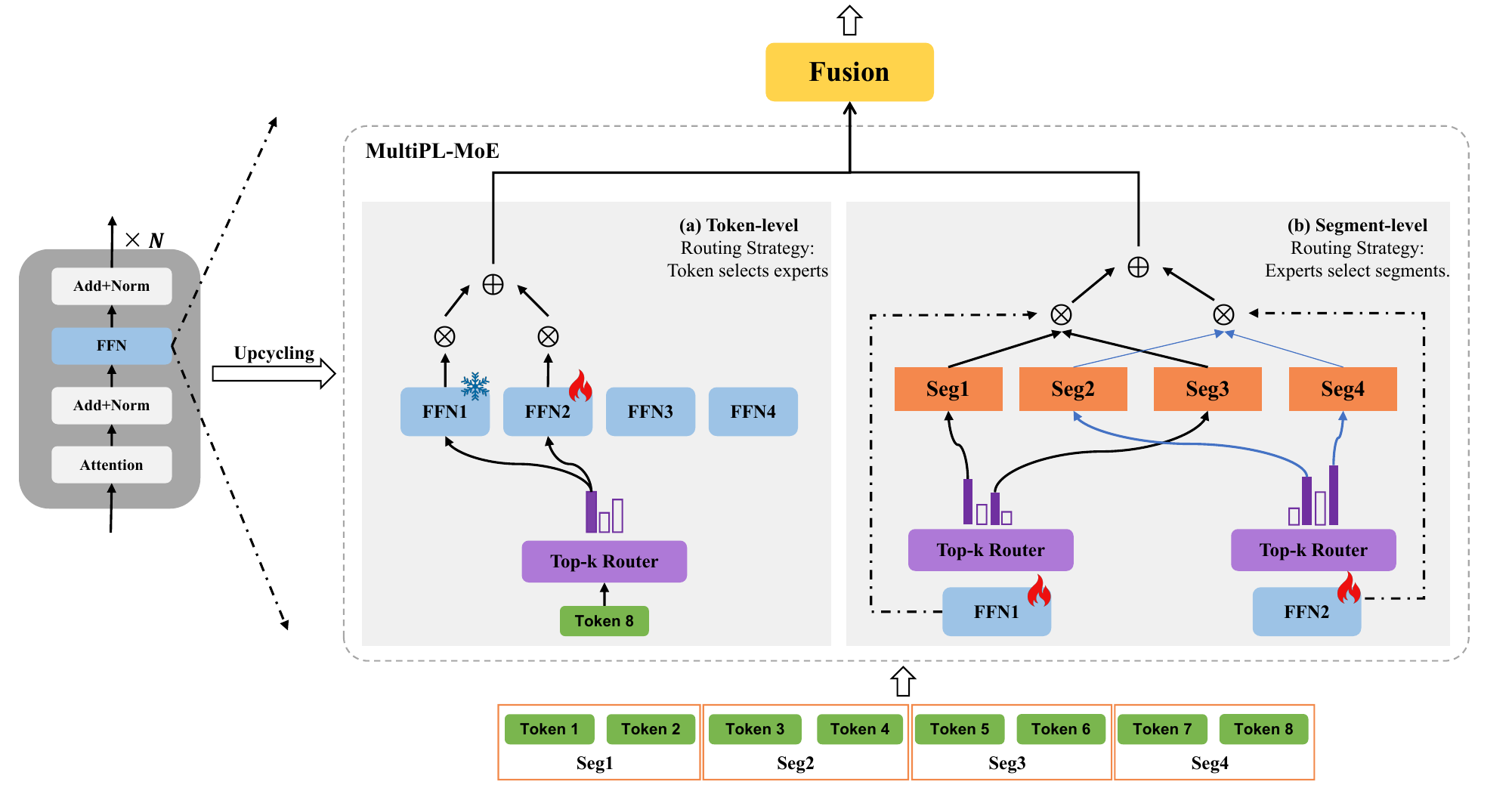}
  \caption{Overall framework of MultiPL-MoE. a)The token-level MoE is a typical MoE structure by upcycling with a shared expert. The routing strategy is token selecting experts. with a novel routing weight normalization method to address scale mismatch.b)The segment-level MoE incorporates two innovative designs to better capture the syntactic structure and contextual patterns of programming languages: 1) Using a sliding window to partition the input token sequence into multiple segments. 2) Adopting a different routing strategy with token-level MoE that allows experts to select the top-k segments. }
  \label{fig:framework}
\end{figure*}

\section{Methodology}
Figure \ref{fig:framework} illustrates the overall framework of MultiPL-MoE, which is constructed through upcycling the original FFN layer of a base LLM. The token-level MoE is a conventional token-choice routing combined with a shared expert \cite{deepseekmoe, ding2024mathcal} with a novel routing weight normalization method to address scale mismatch during the later fusion with the segment-level MoE. For the segment-level MoE, we adopt the expert-choice routing mechanism with the input as contextually coherent segments, enabling experts to capture the syntax structures and discourse-level features. The outputs of the two MoEs are finally fused together.

\subsection{Token-level MoE}
Similar to \citep{deepseekmoe}, the token-level MoE includes a shared expert in each layer to capture common knowledge and reduce redundancy in routed experts. The parameters of the shared expert are fixed during training. The routing strategy is traditional token-choice routing, in which the token selects experts. 

Compared with the standard Transformer, token-level MoE replaces each Feed-Forward Network (FFN) layer with a MoE layer, which uses $N^{tok}$ expert networks that are structurally equivalent to the original FFN layer. A router directs each input token to $K$ out of $N^{tok}$ expert networks. $g_{i,t}$ refers to the gate value for the $i$-th expert given the $t$-th token. Formally, for the $l$-th MoE layer, output hidden state $O_{t,l}^{tok}$  of the $t$-th input token is computed as follows \citep{deepseekmoe}.
\begin{equation}
    \label{eq:vanilla}
    \begin{aligned}
        O_{t,l}^{tok} & = \sum_{i=1}^{N^{tok}}(g_{i,t}FFN_{i}(x_t))  \\
        g_{i,t} & = 
        \begin{cases}
             s_{i,t}, & s_{i,t}\in \mathop{TopK}(S_t,K) \\
             0, & otherwise 
        \end{cases} \\
        S_t & = \{s_{i,t} | 1 \leq i \leq N^{tok}\}  \\
        s_{i,t} & = \mathop{Softmax_i}(x_t \cdot W_{t\_expert})  \\
    \end{aligned}
\end{equation}
where $x_t$ refers to the hidden states of the $t$-th input token for the $l$-th MoE layer. $\mathop{TopK(\cdot)}$ refers to a function extracting K highest scores. $s_{i,t}$ refers
to the affinity score between the $i$-th expert and the
$t$-th token. $W_{t\_expert}$ denotes the router matrix.

However, the scale mismatch between the upcycled MoE layer and the original FFN layer can lead to performance degradation \citep{wu2022residual}. This issue could be even worse for our hybrid MoE architecture because we must finally fuse the outputs of token-level and segment-level MoEs. To facilitate the final fusion, inspired by \cite{feldhus2024}, we enforced the restriction that the sum of gate value $g_{i,t}$ for the selected top-$K$ experts equals 1. Specifically, we first employ the router network to calculate the affinity score of the shared expert for token $t$, denoted as $s_{1,t}$, and select the highest $K-1$ affinity score from the remaining scores. The sum of $K$ affinity scores is then used as a scaling factor $Norm$ to normalize them, ensuring $\sum g_{i,t}=1$. Therefore, the gate value of $g_{i,t}$ can be updated mathematically as follows:
\begin{equation}
    \label{eq:token-level}
    \begin{aligned}
    g_{i,t} & = 
        \begin{cases}
             \frac{1}{Norm}s_{1,t}, & s_{1,t}\in S_t^{'} \\
             \frac{1}{Norm}s_{i,t}, & s_{i,t} \in S_{K,t} \\
             0, & otherwise 
        \end{cases} \\
        S_{K,t} & = \mathop{TopK}(S_t^{'}, K-1) \\
        S_t^{'} & = \{s_{i,t} | 1 < i \leq N^{tok}\}  \\
        s_{i,t} & = \mathop{Softmax_i}(x_t \cdot W_{t\_expert})  \\
        Norm & = \sum S_{K,t} + s_{1,t} \\
    \end{aligned}
\end{equation}
where $S_{K,t}$ denotes the collection of $K-1$ highest router scores among all the experts excluding the shared expert.

\subsection{Segment-level MoE}
Given that experts tend to be underspecialized with token-choice routing \citep{expert-choice}, we hypothesize that such pitfalls may be more severe in learning the segment structures of the programming languages. Therefore, the segment-level MoE adopts the expert-choice routing strategy that independently selects top-$K$ segments for each expert. 

Given a total number of $T$ tokens in an input sample $S=\{token_1, token_2,\dots, token_T\}$, we first partitioned $S$ into $P$ segments based on the context window $a$. These segments are denoted as $SEG=\{seg_1,\dots, seg_p,\dots,seg_P\}$, where $seg_{p} \in \mathbb{R}^{a\times hidden\_size}$, $P=[\frac{T}{a}]$. The expert capacity $r$, which represents the number of segments each expert can take, is defined as follows:
\begin{equation}
\begin{aligned}
    r & =\frac{V\times c}{N^{seg}}, V = \mathcal{B} \times P
\end{aligned}
\end{equation}
where $V$ is the total number of segments within an input batch. Notice that a batch of input includes $\mathcal{B}$ number of samples $S$. $c$ denotes on average how many experts are utilized by a segment, and $N^{seg}$ is the total number of experts for segment-level MoE. 

Three matrices $I$, $D$, and $U$ are employed to produce expert-to-segment assignment. The index matrix $I \in \mathbf{1}^{N^{seg} \times r}$ maps segments to experts, where $I[i,j]$ denotes the $j$-th selected segment by the $i$-th expert. The weight matrix $D \in \mathbb{R}^{N^{seg} \times r}$ assigns expert weight to the selected segment, while $U \in \mathbb{R}^{N^{seg} \times r \times V}$ is an one-hot version of $I$ used to aggregate segments per expert. To facilitate the computation of affinity scores, we derive the embedding $seg_{v}'$ for the $v$-th segment in a batch by averaging the token embeddings within the $v$-th segment. $SEG_{\mathcal{B}}'=\{seg_1',\dots,seg_{v}',
\dots, seg_{V}'\}$ denotes the embeddings of all the $V$ segments within a batch. Following these, the computation procedures of the router and three matrices can be defined as below.
\begin{equation}
    \begin{aligned}
        G_{seg} & = \mathop{Soft\max}(SEG'_{\mathcal{B}} \cdot W_{s\_expert}) \\
        D, I & = \mathop{TopK}(G_{seg}, r), U = \mathop{Onehot}(I)
    \end{aligned}
\end{equation}
where $G_{seg} \in \mathbb{R}^{N^{seg} \times V}$ denotes the expert-to-segment affinity score matrix. $W_{s\_expert}$ denotes the router matrix, and $TopK(\cdot)$ selects the $r$ largest entries for each row of $G_{seg}$. 


Next, we could obtain the input of the $i$-th expert, denoted as $X_{in}[i] \in \mathbb{R}^{r\times hidden\_size}$, by using the matrix $U$ and the segment embeddings for a batch of input samples $SEG'_{\mathcal{B}} \in \mathbb{R}^{V \times hidden\_size}$. The output of the $i$-th expert $X_{out}[i]$ is then generated using the expert's FFN transformation. Given $X_{out}$ and the matrices $U,D$, the output of $l$-th layer at the segment-level $O_{l}^{seg}$ could be calculated as below:
\begin{equation}
    \begin{aligned}
    O_{l}^{seg} & = concat(O_{seg_1},\dots,O_{seg_{v}},...,O_{seg_V})\\
    O_{seg_{v}} & = \sum_{i,j} U[i,j,v]D[i,j]X_{out}[i,j] \\
        X_{out} & = FFN_{i}(X_{in})\\
        X_{in} & = U \cdot SEG'_{batch}
    \end{aligned}
\end{equation}
where $O_{seg_{v}}$ denotes the final output of the $v$-th segment in a batch across all experts.

Finally, we could fuse the $l$-th layer outputs of both the token-level and segment-level MoEs:
\begin{equation}
    Output^l = W^{tok} \cdot O_{t,l}^{tok} + W^{seg} \cdot O_{l}^{seg}
\end{equation}
where $W^{tok}$ and $W^{seg}$ are weight matrices for token-level and segment-level MoEs separately.

\subsection{Training Losses.} 
The hybrid MoE is trained by integrating a next token prediction loss with a load balancing loss, as described below.

\textbf{Next Token Prediction Loss.} Given $T$ tokens, the post-pretraining next token prediction loss is:
\begin{equation}
    L_{NTP} = -\frac{1}{T} \sum_{t=1}^{T} \log p(x_t|Output_{<t})
\end{equation}
where $x_t$ is the true token at position $t$ and $Output_{<t}$ represents the output of all tokens before position $t$.

\textbf{Load Balancing Loss.} Similar to \citep{moe-lpr}, we also incorporate a load balance loss to alleviate the risk of routing collapse in token-level MoE. 
\begin{equation}
\begin{gathered}
    L_{balance}  = \alpha \cdot N^{tok} \cdot \sum_{i=1}^{N^{tok}} f_i \cdot p_i \\
      p_i = \frac{1}{T} \sum_{t \in \mathcal{B}} g_{i,t} \\
    f_i  = \frac{1}{KT} \sum_{t \in \mathcal{B}} \mathbf{1}\{\ Token\,t\,select \, expert \,i\} 
\end{gathered}
\end{equation}
where $\alpha$ is a hyperparameter that controls the weight of the load balancing loss, $K$ indicates the selected experts, $f_i$ denotes the fraction of tokens dispatched to the $i$-th expert and $p_i$ denotes the average affinity scores of tokens allocated to the $i$-th expert. 

The final optimization objective is
\begin{equation}
    \mathop{\arg\min}_{\theta_{*\_expert},W_{*\_expert}} L_{NTP} + L_{balance}
\end{equation}
where $\theta_{*\_expert}$ represents the active experts of both token-level and segment-level, and $W_{*\_expert}$ denotes the router matrix of both token-level and segment-level.

\begin{table*}
 \centering
  \begin{tabular}{lcccccccccc}
    \toprule    
    \multirow{2}{*}{\textbf{Model}} & \multirow{2}{*} {\textbf{$n_{act-params}$}} & \multirow{2}{*}{\textbf{$n_{params}$}} & 
     \multicolumn{6}{c} {\textbf{Programming Languages}} & \multirow{2}{*}{\textbf{Avg.}} &  \\
    \cline{4-9}
    & & & \textbf{Python}  & \textbf{Java}  & \textbf{C++}  & \textbf{Rust}  & \textbf{Go} & \textbf{Ruby} \\
    \hline
    \textbf{HumanEval} & & & & & & & \\ 
    \hline
     {Qwen1.5} & {1.8B} & {1.8B} & \underline{{20.1}} & {10.3} & \underline{{10.0}} & {0.5} & {2.3} & {1.1} & {7.4} \\
     {Post-pretraining} & {1.8B} & {1.8B}  & \textbf{{22.8}} & \underline{{10.4}} & \underline{{10.0}} & {0.9} & {2.3} & {0.6} & {7.8} \\
     {LoRA} & {1.8B} & {1.8B} & \underline{{20.1}} & {7.2} & {5.7} & {0.1} & {2.1} & {0.1} & {5.9} \\ 
     {Gshard} & {3.5B} & {10.8B} &  {7.9} & {2.1} & {7.9} & \underline{{7.3}} & {7.4} & {6.0} & {6.4} \\
     {Mixtral} & {2.6B} & {7.5B} & {8.5} & {1.6} & {2.1} & {0.7} & {2.9} & {4.6} & {3.4} \\
     {Qwen1.5-MoE} & {2.7B} & {14.3B} & {15.2} & {9.2} & {8.6} & {6.1} & \underline{{8.2}} & \underline{{6.5}} & \underline{{9.0}} \\
     {\textbf{MultiPL-MoE}} & {3.5B} & {10.8B} &  {16.6} & \textbf{{10.6}}	& \textbf{{10.6}} & \textbf{{10.9}} & \textbf{{8.7}} & \textbf{{7.1}} & \textbf{{10.8}} \\
    \hline
    \textbf{MBPP} & & & & & & & & & \\ 
    \hline
    {Qwen1.5} & {1.8B} & {1.8B} & \textbf{{18.0}} & \underline{{19.7}}	& \underline{{19.4}} & {4.5} & {8.5} & {0.2}  & {10.0} \\
    {Post-pretraining} & {1.8B} & {1.8B} & \underline{{4.6}} & \textbf{{20.4}} & \textbf{{19.6}} & {5.1} & {9.1} & {0.2} & {9.8} \\
    {LoRA} & {1.8B} & {1.8B} & {2.4} & {12.1} & {16.9} & {0.9} & {6.9} & {0.3} & {6.6} \\
    {Gshard} & {3.5B} & {10.8B} &  {0.2} & {9.0} & {11.6} & {11.2} & {12.6} & \underline{{16.4}} & {10.2} \\
    {Mixtral} & {2.6B} & {7.5B} & {0.0} & {9.1} & {9.6} & {5.5} & {10.7} & {3.9} & {6.5} \\
    {Qwen1.5-MoE} & {2.7B} & {14.3B} & {0.0} & {17.6} & {18.0} & \textbf{{17.1}} & \underline{{17.1}} & {16.0} & \underline{{14.3}} \\
    {\textbf{MultiPL-MoE}} & {3.5B} & {10.8B} &  {1.4} &	{17.3}	& {18.0} & \underline{{16.1}} & \textbf{{17.3}} & \textbf{{19.6}} & \textbf{{15.0}} \\
    \bottomrule
  \end{tabular}
  \caption{Evaluation results across six programming languages on HumanEval and MBPP. Python, Java, C++ are high-resource programming languages. Rust, Go and Ruby are low-resource programming languages. $n_{\text{act-params}}$ is the number of activated model parameters per token, and $n_{\text{params}}$ is the total number of model parameters. \textbf{Bold} denotes the best, and \underline{underline} denotes the second-best.}
  \label{tab:main-results}
\end{table*}

\section{Experiments}
\subsection{Experiment Setup}
\textbf{Datasets.} We select Qwen1.5-1.8B \citep{qwen1.5} as our backbone model due to its lower computational requirements and ease of upcycling. Our focus is on three low-resource programming languages: Rust, Go, and Ruby, for which Qwen1.5-1.8B exhibits poor performance (see Figure \ref{fig:code-polar}). Additionally, we include Python, Java, and C++ as high-resource programming languages to examine their catastrophic forgetting during model updates. We sample 3.3 billion tokens from The Stack \citep{The-Stack} for post-pretraining, with a 9:1 ratio of low- and high-resource programming languages.

\textbf{Implementation Details.} MultiPL-MoE upcycles Qwen1.5-1.8B with 12 experts per layer, integrating 6 token-level experts (including a shared expert) with top-2 activation and 6 segment-level experts configured as $c=1$, $r=10$. During post-pretraining, the upcycled MoE model updates only the newly introduced experts and the routing mechanism. The training settings include a batch size of 64, a sequence length of 1024, a learning rate of 7e-6 with a cosine scheduler, and an $\alpha$ of load balancing loss of 0.01. We  utilize BF16 mixed precision and flash attention \citep{flashattention} to speed up the training process. We evaluate on HumanEval \citep{HumanEval} and MBPP \citep{MBPP} benchmarks extended to multiple programming languages through MultiPL-E adaptation \citep{MultiPL-E}.

\textbf{Baselines.} We conduct experiments on several existing baselines on the same dataset to evaluate the efficiency of MultiPL-MoE. 
\begin{itemize}[noitemsep,topsep=0pt]
    \item \textbf{Post-pretraining}: Directly perform continue training on Qwen1.5-1.8B.
    \item \textbf{LoRA} \citep{lora}: LoRA settings are applied to all linear layers, and the rank is 8.
    \item \textbf{Gshard} \citep{gshard}: In GShard, we set up 12 experts and select the top 3 routing strategy.
    \item \textbf{Mixtral} \citep{mixtral}: Each layer employs 8 experts, with per-token routing dynamically selecting and aggregating outputs from the top 2 experts.
\item \textbf{Qwen1.5-MoE} \citep{qwen_moe}: We upcycle the Qwen1.5-1.8B model to a Qwen-MoE structure with 2.7B activated parameters. 
\end{itemize}

\subsection{Experiment Results}
\textbf{Main results and analysis.} Table \ref{tab:main-results} summarizes the pass@1 results of baselines on both low-resource (Rust, Go, Ruby) and high-resource (Python, Java, C++) programming languages under the Humaneval and MBPP benchmarks. The results demonstrate that MultiPL-MoE achieves significant advancements in cross-language generalization, particularly for low-resource programming languages. Compared to MoE architectures, the dense models (Post-pretraining and LoRA) show superior performance on high-resource programming languages across both evaluation benchmarks, while exhibiting degraded effectiveness on low-resource programming languages. MultiPL-MoE achieves 10.8 pass@1 on HumanEval and 15.0 pass@1 on MBPP with 3.5B activated parameters, greatly surpassing Gshard. MultiPL-MoE and Qwen1.5-MoE achieve consistent performance across both benchmarks, significantly enhancing model capabilities on low-resource programming languages while effectively mitigating catastrophic forgetting in high-resource programming languages.

The results further reveal divergent patterns across benchmarks. MultiPL-MoE achieves state-of-the-art average performance on HumanEval, surpassing the second-best Qwen1.5-MoE by 1.8 points. Notably, our model dominates all programming languages except Python, with Ruby exhibiting the greatest improvement. Although Qwen1.5-MoE is competitive, our model maintains its lead with a higher average pass@1 on MBPP. Post-pretraining performs competitively on MBPP in C++ and Java, but falls short in low-resource languages. We notice that Python exhibits consistently low performance on MBPP across all models except the base model, which is substantially inferior to other languages.  

\textbf{Effect of Key Components.} We conduct experiments to evaluate the impact of two critical components in MultiPL-MoE: (1) segment-level MoE and (2) shared expert mechanisms at the token-level. Experiments are carried out using the HumanEval benchmark, comparing pass@1 scores for high- and low-resource programming languages. 

\begin{table}[htbp]
  \centering 
  \adjustbox{max width=\columnwidth}{
  \begin{tabular}{lcccc}
    \toprule
    \multirow{2}{*}{\textbf{Model}} & \multicolumn{2}{c}{\textbf{HumanEval}}  \\
    \cline{2-3}
    & \textbf{Original} & \textbf{Expanded} &  \\
    \hline
    Qwen1.5     & {40.4} & {3.9}\\
    \hline
    {MultiPL-MoE}  & \multicolumn{3}{c}{} \\
     \hspace{2em} \small -w/o segment-level  & {} & {} \\
     \hspace{2em} \small -w/o shared expert &
    \multirow{-3}{*}{27.3} &  \multirow{-3}{*}{22.5}  \\
    \hline
    {MultiPL-MoE}  & \multicolumn{3}{c}{} \\  
    \hspace{2em} \small -w/o segment-level &
    \multirow{-2}{*}{30.9} &  \multirow{-2}{*}{16.8}  \\
    \hline
    {MultiPL-MoE}  & \multicolumn{3}{c}{} \\  
    \hspace{2em} \small -w/o shared-expert &
    \multirow{-2}{*}{33.3} &  \multirow{-2}{*}{22.3}  \\
    \hline
    MultiPL-MoE  & \textbf{{37.8}} & \textbf{{26.7}}  \\
    \bottomrule
  \end{tabular}
  }
  \caption{Performance evaluation on HumanEval, comparing the aggregated pass@1 scores between high-resource (Original) and low-resource (Expanded) programming languages. Dual exclusion (w/o) means the absence of both configurations.}
  \label{table:ablation}
\end{table}

\begin{figure*}[t]
 \includegraphics[width=\textwidth, height=4cm]{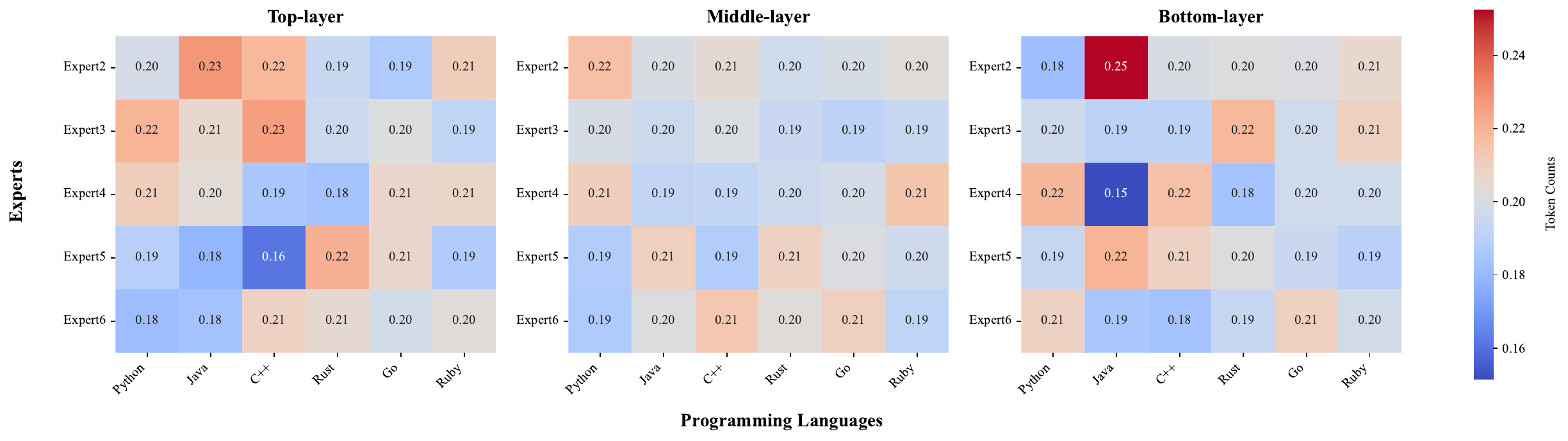}
  \caption{Frequency of selected tokens in different layers. The shared expert (first expert) is omitted from schematic diagram due to its universal application. Our 24-layer architecture spans layer 1 (bottom) to layer 24 (top) with layer 12 as the midpoint.}
  \label{fig:token-visual}
\end{figure*}

As shown in Table \ref{table:ablation}, the complete MultiPL-MoE achieves state-of-the-art performance, outperforming the Qwen1.5 on low-resource programming languages by 22.8 points, while exhibiting minor degradation on high-resource programming languages. This trade-off illustrates MultiPL-MoE's capacity to favor cross-lingual generalization while avoiding catastrophic forgetting. Dual exclusion of both components causes severe degradation in high-resource programming languages. The removal of segment-level MoE affects performance for low-resource programming languages, demonstrating its utility. In contrast, the deletion of shared experts has a significant impact on adaptation to high-resource languages, resulting in lower pass@1 scores. These results demonstrate that segment-level MoE is essential for capturing language-specific syntactic patterns in low-resource languages, whereas shared experts stabilize cross-lingual knowledge transfer and mitigate catastrophic forgetting.

\subsection{Expert Specialization Analysis}

\begin{figure*}[t]
  \includegraphics[width=\textwidth,height=6.5cm]{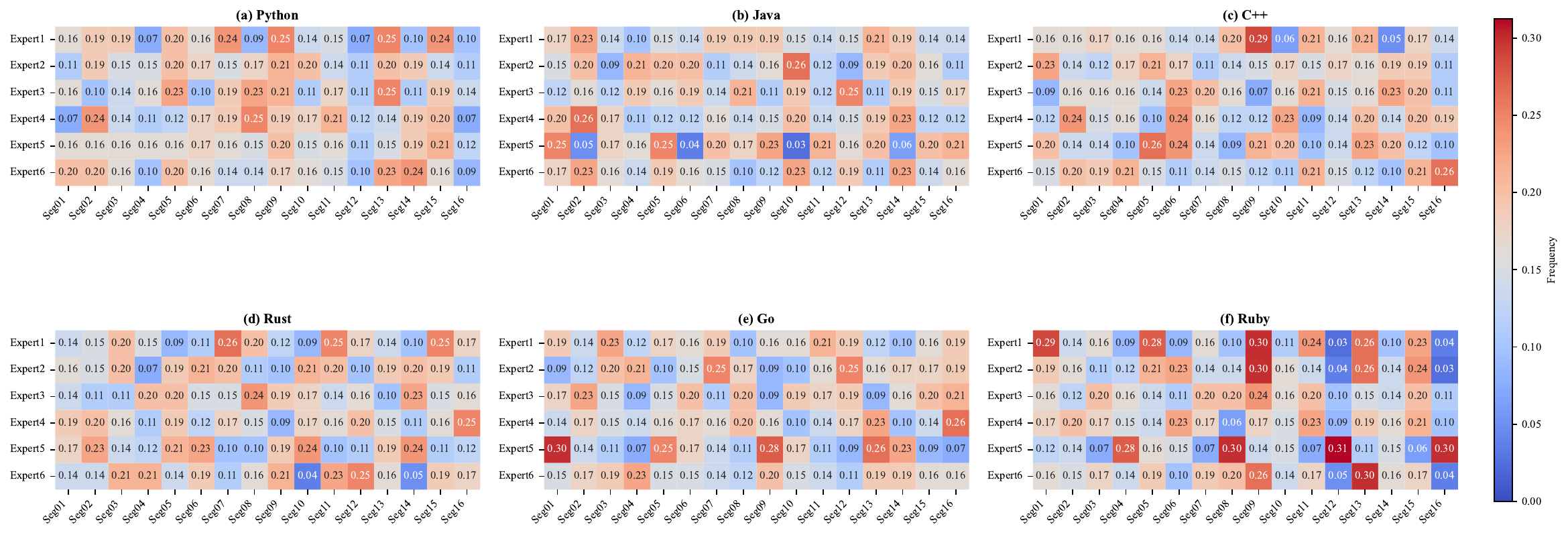}
  \caption{Frequency of selection by segment-level experts in the different programming languages at the top layer (Layer 24) of MoE. Experts at the segment level prefer different segments across six programming languages.}
  \label{fig:seg-visual}
\end{figure*}

To investigate the specialization patterns of hybrid MoE experts, we examine the selection frequency of experts at the token and segment levels in six programming languages. We compare the frequency with which experts are selected in the top layer (layer 24), middle layer (layer 12), and bottom layer (layer 1) to trace the hierarchical transmission of specialization.

\textbf{Token-level experts specialization.} Figure \ref{fig:token-visual} illustrates the progressive attention to token-level experts over different MoE layers. Specialization patterns emerge among bottom-layer experts, with distinct programming language preferences. Quantitative analysis shows that Expert 2 demonstrates peak Java competency, while Expert 3 shows Rust and Ruby proficiency. Expert 4 maintains balanced Python and C++ engagement, and Expert 6 concentrates exclusively on Python and Go. Middle-layer specialists, on the other hand, display transitional traits, with values clustering primarily around 0.20.

Experts in the top-layer also exhibit distinct specialization patterns divergent from bottom-layer experts, demonstrating the architecture’s capacity to aggregate abstract cross-linguistic features. Specifically, Expert 5 specializes in procedural programming paradigms (Rust and Go), while Expert 2 and Expert 3 focus predominantly on object-oriented programming languages. Expert 4 and 6 exhibit no statistically significant linguistic preferences. Notably, Ruby and Go's constantly balanced expert attention across all layers highlights the ongoing difficulty of representing low-resource programming languages, which may result from limited cross-language interchangeability. 

\begin{table}[htbp]
  \centering 
  \adjustbox{max width=\columnwidth}{
  \begin{tabular}{lcccc}
    \toprule
    \multirow{2}{*}{\textbf{Language}} & \multicolumn{3}{c}{\textbf{Top-2 Selected Segments}}  \\
    \cline{2-4}
    & \textbf{L1} & \textbf{L12} &  \textbf{L24} \\
    \hline
    Python    & {(10, 3)} &  {(3, 11)} &  {(9, 13)} \\
    Java     &  {(12, 5)} & {(5, 15)} & {(2, 14)} \\
    C++     &  {(14, 10)} & {(10, 14)} & {(6, 15)}  \\
    Rust     & {(5, 2)} & {(6, 2)} & {(11, 15)}\\
    Go     &  {(1, 5)} & {(1, 5)} & {(16,13)}  \\
    Ruby    &  {(8, 4)} & {(8,4)} & {(9,13)}\\
    \bottomrule
  \end{tabular}
  }
  \caption{Top-2 selected segments at three layers across six programming languages. ((10,3) represents the segment 10 and 3, which are the most frequently selected by segment-level experts at the corresponding layer. L1 denotes the bottom-layer (layer 1).}
  \label{table:seg}
\end{table}

\textbf{Segment-level experts specialization.} We investigate segment selection frequencies from segment-level experts across six programming languages, focusing on how language-specific and structural preferences impact experts' specialization. Table \ref{table:seg} demonstrates distinct segment selection patterns. Specially, layer 1 and layer 12 experts exhibit identical top-2 segment activation patterns for C++, Go and Ruby, suggesting cross-linguistic consistency in low-level feature processing. Experts in layer 24 exhibit divergent segment preferences.  

Figure \ref{fig:seg-visual} illustrates that segment-level experts at the top-layer (layer 24) show varied preferences for segments. Experts at the top layer primarily concentrate on segments 5, 9, 12, and 16 across six programming languages, exhibiting cross-linguistic consistency. In particular, Experts appear to prefer segments 5 and 9 in high-resource programming languages (Python, Java, and C++) while engaging more with segments 12 and 16 in low-resource programming languages (Rust, Go, and Ruby). We observe that expert 5 shows distinct preference patterns in different programming languages, excluding Python. 

To conduct comparative analysis, we also analyze the segment selection frequencies of experts in layers 1 and 12, see Appendix \ref{sec:seg}.



\begin{figure}[t]
  \includegraphics[width=\columnwidth,height=4.5cm]{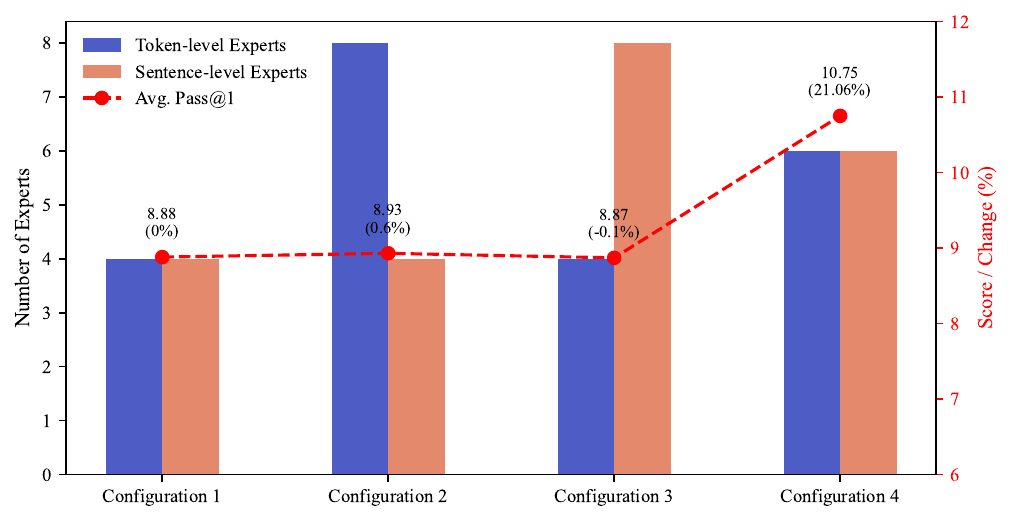}
  \caption{Average performance results for different numbers of experts on HumanEval. The red dashed line denotes the average pass@1 score, computed by averaging pass@1 scores across six programming languages.}
  \label{fig:alloc_expert}
\end{figure}

\textbf{Allocation of Experts.} 
We analyze the relationship between expert configurations and multilingual code reasoning performance. Due to MultiPL-MoE's hybrid two-level MoE design, we investigate four different configurations with varied token-level to segment-level expert ratios (4:4, 8:4, 4:8, and 6:6). These configurations are subsequently referred to as Configuration 1 through 4 in Figure \ref{fig:alloc_expert} for comparative performance analysis. 

As illustrated in Figure \ref{fig:alloc_expert}, Configuration 1 yields a baseline average pass@1 score of 8.88. Expanding token-level expertise while maintaining segment-level capacity of Configuration 2 yields marginal improvement to 8.93, whereas the inverse allocation of Configuration 3 results in slight degradation to 8.87. Notably, the balanced allocation of Configuration 4 achieves the highest average pass@1 score of 10.74, surpassing the Configuration 1 score of 20.91\%. The superior performance of Configuration 4 conclusively demonstrates that balanced allocation across two levels is essential for optimal multilingual code reasoning. 

\section{Conclusion}
This paper proposes MultiPL-MoE, a hybrid MoE with both token-level and segment-level. MultiPL-MoE introduces shared experts to capture the commonality of knowledge at the token level, and obtains semantic and logical information between segments at the segment level. Extensive empirical results on two different benchmarks demonstrate the effectiveness of MultiPL-MoE. In summary, MultiPL-MoE is an effective method for extending low-source programming languages in the post-pretraining phase.

\section{Limitations}
To capture the syntactic structure and contextual patterns of programming languages, we introduce a segment-level MoE with a fixed sliding window mechanism. It would be interesting to explore a dynamic sliding window strategy based on semantic or syntactic code features. Furthermore, while our empirical results validate the effectiveness of MultiPL-MoE, it would be interesting to provide a complete theoretical explanation for the hybrid MoE architecture.

\bibliography{custom}

\appendix

\section{Appendix}
\subsection{Comparison of segment frequencies at different MoE layers in six programming languages} \label{sec:seg}

\begin{figure*}[t]
  \includegraphics[width=\textwidth]{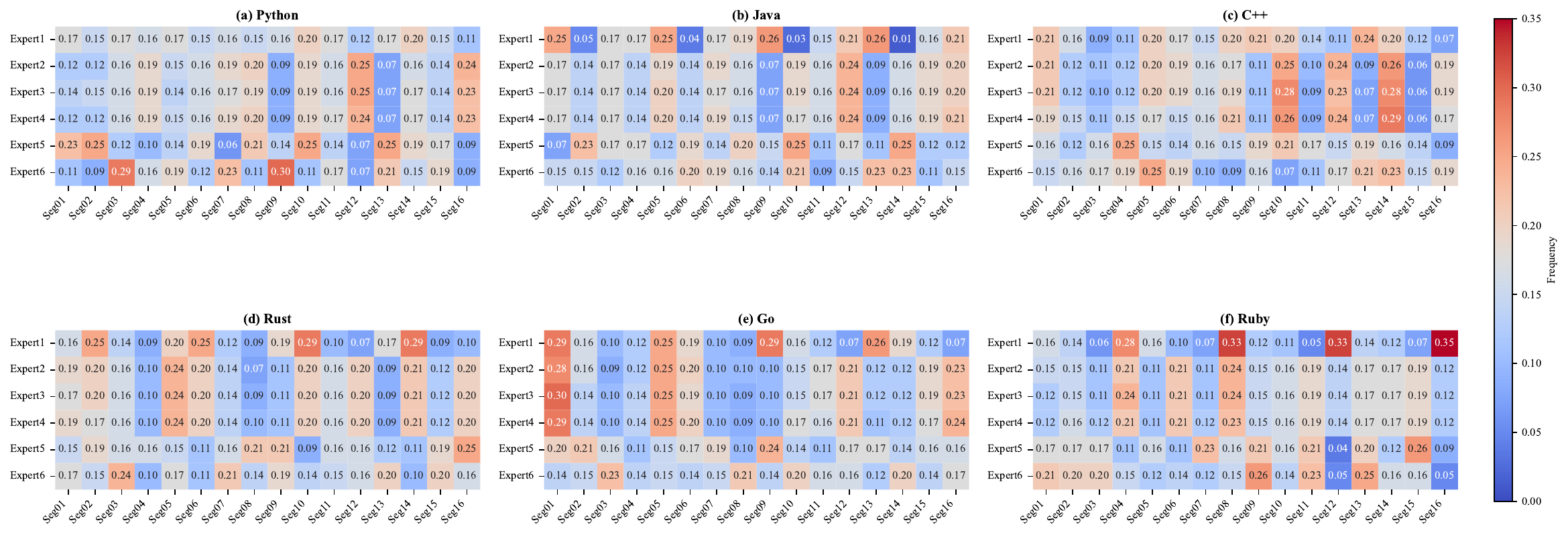}
  \caption{At the bottom-layer of the MoE layer, experts in specific fields have their own unique preferences for different sub-fields of the six programming languages.} 
  \label{fig:seg-visual-layer1}
\end{figure*}

\begin{figure*}
  \includegraphics[width=\textwidth]{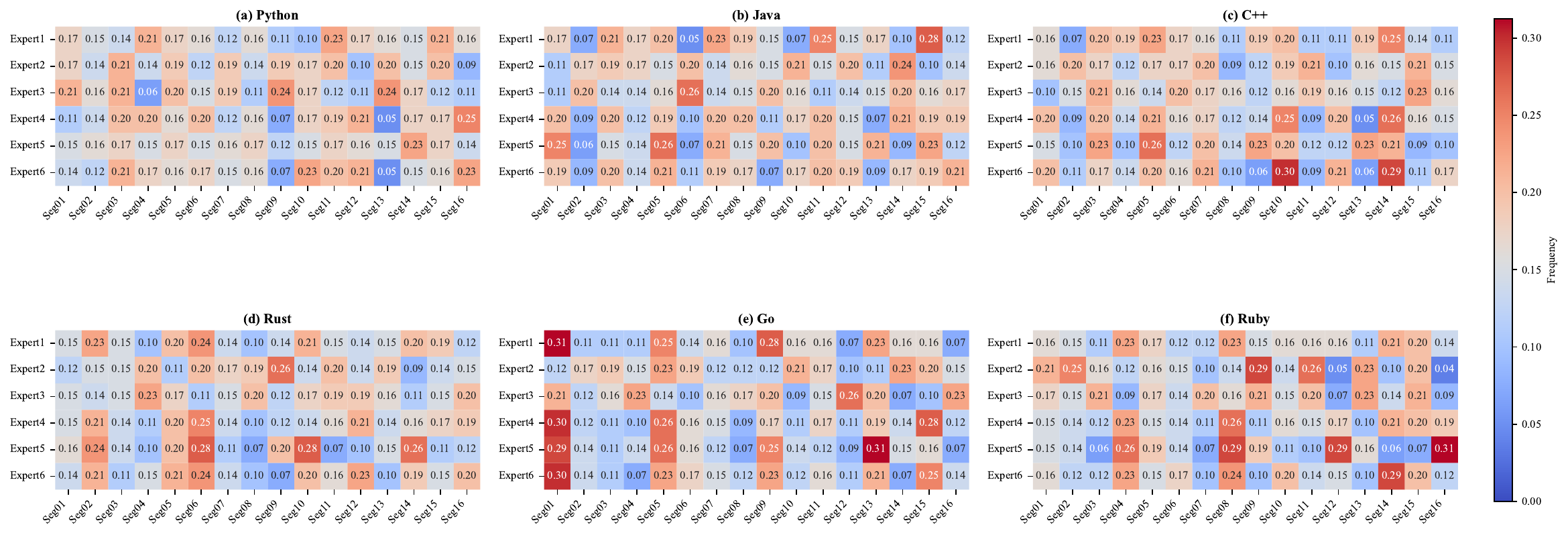}
  \caption{At the middle-layer of the MoE layer, experts in niche areas have their own unique preferences for different niches of six programming languages.} 
  \label{fig:seg-visual-layer12}
\end{figure*}

By comparing Figure \ref{fig:seg-visual-layer1}, Figure \ref{fig:seg-visual-layer12}, and Figure \ref{fig:seg-visual}, we conduct comparative analysis from the following two perspectives. From the perspective of model depth, at the shallow level, the segment selections of experts in a single programming language show a high degree of convergence, indicating their universal dependence on the basic grammatical structure. As the number of model layers increases, the expert selection strategies gradually diverge, reflecting the gradual emergence of expert division of labor. From the perspective of cross-language comparison, the segment selections of the same expert in different programming languages vary significantly, indicating that expert decision-making is language-specific. This difference may come from the essential differences between different programming languages in grammatical rules, semantic expressions, and code idioms.

\end{document}